\documentclass[10pt,twocolumn,letterpaper]{article}

\usepackage[pagenumbers]{cvpr}      %

\usepackage[dvipsnames]{xcolor}

\definecolor{cvprblue}{rgb}{0.21,0.49,0.74}
\usepackage[pagebackref,breaklinks,colorlinks,citecolor=cvprblue]{hyperref}
\usepackage{epigraph} 
\usepackage[utf8]{inputenc}
\usepackage{csquotes}
\usepackage{wrapfig}
\usepackage{float}
\usepackage{bm}

\usepackage[capitalize]{cleveref}
\usepackage{graphicx}
\usepackage{xspace}
\usepackage{booktabs}

\definecolor{MyDarkGreen}{rgb}{0.02,0.6,0.02}

\newcommand{\method}{\textsc{Chord}\xspace}

\crefname{section}{Sec.}{Secs.}
\crefname{table}{Table}{Tables}
\crefname{figure}{Figure}{Figures}

\usepackage{multirow}
\usepackage{makecell}

\newcommand{\myparagraph}[1]{\vspace{0.1cm}\noindent\textbf{#1}}

\title{Choreographing a World of Dynamic Objects}

\author{}

\author{
Yanzhe Lyu$^{1,*,\text{†}}$
\qquad
Chen Geng$^{1,*}$
\qquad
Karthik Dharmarajan$^{1}$ \\
\qquad
Yunzhi Zhang$^{1}$ 
\qquad
Hadi Alzayer$^{1,3}$
\qquad
Shangzhe Wu$^{2}$
\qquad
Jiajun Wu$^{1}$ \\[0.8em]
$^1$Stanford University \qquad
$^2$University of Cambridge  \qquad
$^3$University of Maryland
}

\begin{document}

\twocolumn[{
    \renewcommand\twocolumn[1][]{#1}
    \maketitle
    \vspace{-3.0mm}
    \begin{center}
    \vspace{-3.0mm}
        \captionsetup{type=figure}
        \includegraphics[width=\textwidth]{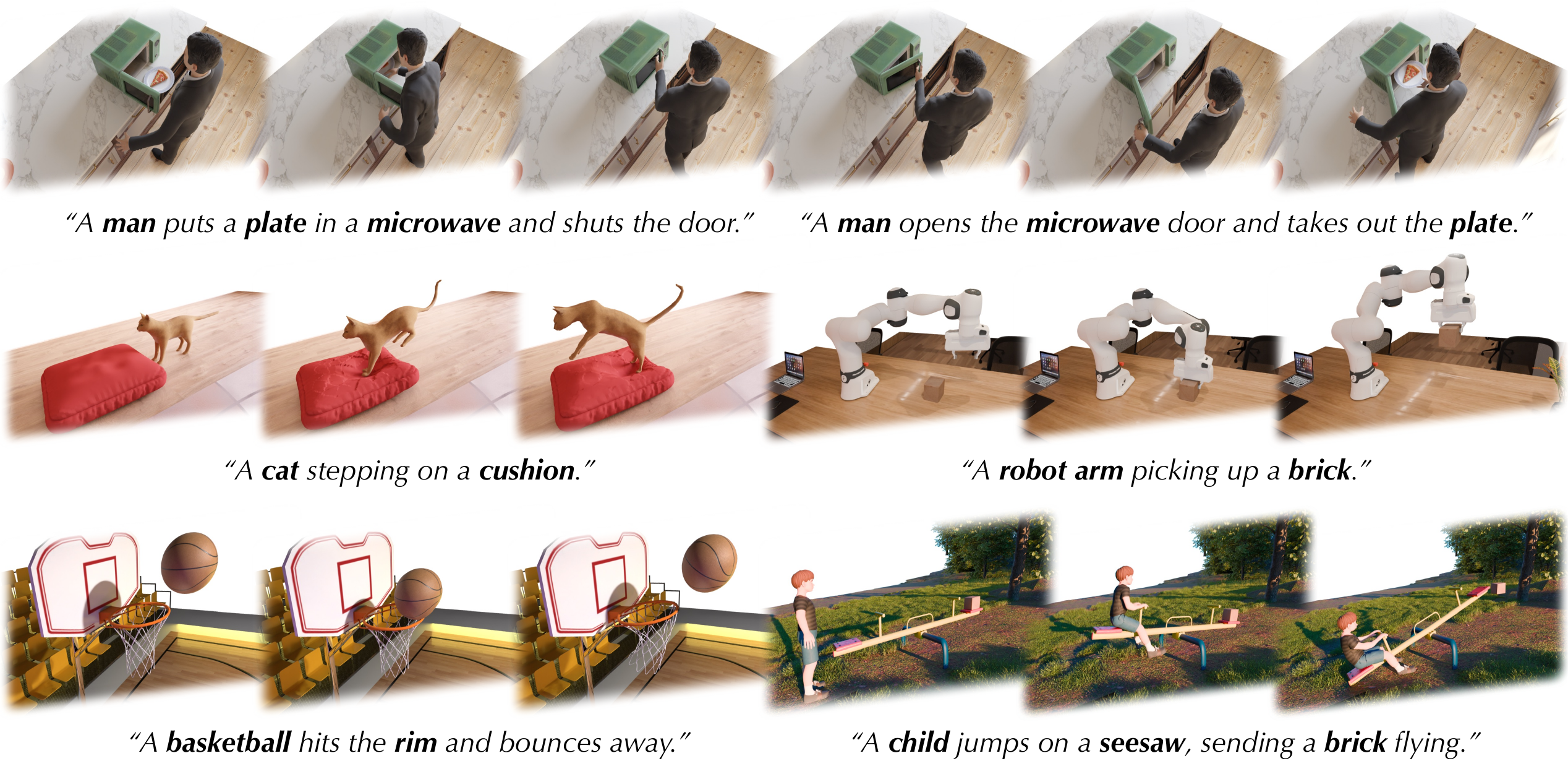}
        \captionof{figure}{
        \textbf{4D scene motion generated by our method.} We present \method, a universal generative pipeline capable of animating scenes with \textbf{multiple} objects that interact with each other. Project page: \url{https://yanzhelyu.github.io/chord}   
        }
        \label{fig:teaser}
    \end{center}
    }
    \bigbreak
    ]

{
\let\thefootnote\relax\footnotetext{$^*$Equal contribution. $^{\text{†}}$Work was done when Y. Lyu was a visiting student at Stanford University. Y. Lyu is currently with the University of Science and Technology of China.}
}
\begin{abstract}

Dynamic objects in our physical 4D (3D + time) world are constantly evolving, deforming, and interacting with other objects, leading to diverse 4D scene dynamics. In this paper, we present a universal generative pipeline, \method, for \textbf{CHOR}eographing \textbf{D}ynamic objects and scenes and synthesizing this type of phenomena. Traditional rule-based graphics pipelines to create these dynamics are based on category-specific heuristics, yet are labor-intensive and not scalable. Recent learning-based methods typically demand large-scale datasets, which may not cover all object categories in interest. Our approach instead inherits the universality from the video generative models by proposing a distillation-based pipeline to extract the rich Lagrangian motion information hidden in the Eulerian representations of 2D videos. Our method is universal, versatile, and category-agnostic. We demonstrate its effectiveness by conducting experiments to generate a diverse range of multi-body 4D dynamics, show its advantage compared to existing methods, and demonstrate its applicability in generating robotics manipulation policies. Project page: \url{https://yanzhelyu.github.io/chord}

\end{abstract}
    
\section{Introduction}
\label{sec:intro}

Humans and other embodied agents live in a 4D (3D + time) world, a world composed of a diverse range of dynamic objects, \ie, objects that can evolve, deform, or interact with other objects. Creating 4D motions for both object deformations and interactions is crucial when building 3D world models for robotics~\cite{mandi2025dexmachina,xiang2020sapien} and embodied AI~\cite{li2023behavior}.

Traditionally, it has been challenging to generate such motions for a scene composed of dynamic objects in their static snapshots because it requires extensive manual modeling and expert labor. 
Recent approaches~\cite{wu2025animateanymesh} have attempted to learn such 4D generators purely from data in an end-to-end manner. However, most existing datasets~\cite{deitke2023objaverse} focus on the internal deformations and evolutions of an individual object with little to no coverage on their interactions, and 4D data describing both deformations of objects and object interactions is extremely rare. This scarcity on scene-level 4D dynamics has rendered existing data-driven approaches into only being capable of generating dynamics of a single object.

Inspired by the recent success of general-purpose video generative models, we use a different approach to tackle this problem: distilling these scene motions from video generative models. At a high level, we iteratively optimize the low-level Lagrangian deformations of each object. At each optimization step, we deform the 3D scene and render it from certain viewpoints, and let video generative models judge whether the deformation is plausible. Through this process, we essentially leverage video models as a high-level ``choreographer'' to plan the motions of individual objects and make them consistent with each other.

Despite the promise of this distillation-based paradigm, getting plausible results with it has been challenging. Existing methods~\cite{bahmani20244d,jiang2024animate3d,uzolas2025motiondreamer} mainly operate at the object level and often show noticeable artifacts in the generated motion. Two major obstacles hinder these approaches from working effectively in our setting: (1) 4D deformations are both spatially high-dimensional and temporally ill-regularized, and (2) the non-conventional architecture designs of modern video generative models are not compatible with existing distillation algorithms~\cite{poole2023dreamfusion}. 

We address the first challenge by analyzing the inherent locality of 4D deformations: temporal deformation fields should be locally smooth in both space and time. To this end, we design a coarse-to-fine 4D motion representation that injects hierarchical structures to both the spatial and the temporal domain. Spatially, we adopt a bi-level control point-based representation that disentangles fine-grained motion details from coarse transformations. Temporally, inspired by a time-honored data structure in theoretical algorithm design, i.e. the Fenwick tree~\cite{kulal2021hierarchical,donald1999taocp}, we store deformations in a cumulative, range-based structure that implicitly enforces temporal coherence and improves the learnability of long-horizon motion. With these two innovations, our novel 4D representation is robust, stable, and supports generating a diverse range of motions.

The second challenge stems from modern video generative models being based on flow-based models~\cite{liu2023flow}. These models are incompatible with the traditional distillation algorithms. Therefore, we propose a novel strategy for distillation from modern rectified flow-based video generative models. We derive a novel Score Distillation Sampling (SDS) \cite{poole2023dreamfusion} target for flow-based video diffusion models and analyze their noise pattern, thus enabling video models to effectively provide guidance to our 4D representation.

By proposing these two innovations and the framework to choreograph object motion, we arrive at a simple yet elegant solution to the challenging problem of generating 4D-consistent motion of dynamic objects in a scene. We name this pipeline ``\method'', for \textbf{CHOR}eographing \textbf{D}ynamic objects and scenes. \method is universal, versatile, and applicable across a wide range of dynamic phenomena. We evaluate our framework on diverse dynamic objects and compare it against prior art and show clear advantages. 

Beyond visual generation, our pipeline also enables the robot manipulation in the physical world by generating physically-grounded Lagrangian deformation trajectories of real-world objects. We demonstrate this by leveraging the generated 3D trajectories to plan the motion of a real robot and showing that they can guide zero-shot manipulation of diverse dynamic objects. 

In summary, our contributions are as follows:

\begin{enumerate}
    \item A 4D motion representation that combines a Fenwick tree–inspired cumulative temporal structure with a hierarchical low-to-high DoF parameterization, making it well-suited for distillation-based 4D generation.
    \item A distillation strategy for modern flow-based video generative models to make SDS algorithms effective on generating 4D motions from 2D video generative models.
    \item A robust framework to generate physically-grounded 4D motions for diverse dynamic objects that are applied to learning real-world robotic manipulation policies.
\end{enumerate}

\section{Related Work}
\label{sec:related}

\myparagraph{Object-Level 4D Generation.} Generating 4D consistent object deformations has been a long-standing challenge in the community. Traditional approaches first determine category-specific kinematic models (\ie, rigging representations)~\cite{loper2023smpl,pavlakos2019expressive,blanz2003face,booth20163d,Zuffi:CVPR:2017,wu2023magicpony,he2025canor,liu2024differentiable} and then generate motion based on them~\cite{huang2025animax,mahmood2019amass,peng2023implicit,geng2023learning,su2021nerf,xu2024relightable,liu2021neural,tevet2022human,song2025puppeteer}, which inherently limits these methods to constrained categories. Some methods~\cite{zhang2025gaussian,wu2025animateanymesh,yenphraphai2025shapegen4d} attempt to learn end-to-end 4D generators from existing 4D object datasets~\cite{deitke2023objaverse,deitke2023objaversepp,deng2025anymate}, but they struggle to generalize beyond humanoid characters since most existing datasets are dominated by animated human-like models. Other approaches~\cite{bahmani20244d,gao2024gaussianflow,jiang2023consistent4d,li2024dreammesh4d,liang2024diffusion4d,ren2023dreamgaussian4d,ren2024l4gm,sun2024eg4d,uzolas2025motiondreamer,wang2024vidu4d,xie2024sv4d,yang2024diffusion,yu20244real,yuan20244dynamic,geng2025birth,yao2025sv4d,mou2025dimo,ma20254d} avoid supervised learning by performing 4D reconstruction or distillation from video generative models, yet they typically yield minor and unrealistic motion due to the difficulty of optimizing high-dimensional 4D motion and the noise in the guidance signals. Our framework addresses these limitations by assuming neither category-specific kinematic structure nor large-scale 4D datasets, and generates realistic 4D motion for arbitrary objects.

\myparagraph{Scene-Level 4D Generation.} Scene-level 4D generation extends beyond the object-centric setting, introducing substantially more complexity and greater challenges. It must not only produce plausible object-level motion but also maintain motion consistency across multiple interacting objects. Therefore, existing methods often simplify the problem by restricting it to specific categories (\eg, human-object interaction~\cite{ye2024g,huang2022reconstructing,wen2023bundlesdf,li2024zerohsi}), enforcing physical constraints~\cite{zhang2024physdreamer,chen2025physgen3d,lin2025omniphysgs,li2025wonderplay}, or conditioning on symbolic structures~\cite{bahmani2024tc4d,shuai2022novel}. Some approaches attempt to produce 4D scenes by reconstructing them from videos ~\cite{chu2024dreamscene4d,wu2025cat4d,lei2025mosca,wang2025shape} generated by video models, yet the resulting representation remains largely 2.5D and does not support full 360$^\circ$ view synthesis. Our approach is the first to tackle the challenging setting of generating scene-level 4D motion of objects without relying on any category-specific inductive bias.

\myparagraph{4D Representations.} A key component in 4D generation pipelines is the selection of the underlying 4D representation. Early works use high-dimensional deformation fields to represent 4D scenes~\cite{wu20244d,gao2024gaussianflow,park2021nerfies,pumarola2021d}. They work well for reconstruction targets with dense inputs, but are not suitable for generative tasks with noisy supervision signals. Recent works explore reducing the dimensionality of 4D representations in the spatial domain~\cite{wu2024sc4d,he2025canor,wang2025shape,geng2025birth}. Our hierarchical 4D representation strengthens this idea by injecting low-dimensionalities and hierarchies in both spatial and temporal domains, which serves as a backbone representation in our 4D generation framework.

\section{Method}

\begin{figure*}[t]
\includegraphics[width=\linewidth]{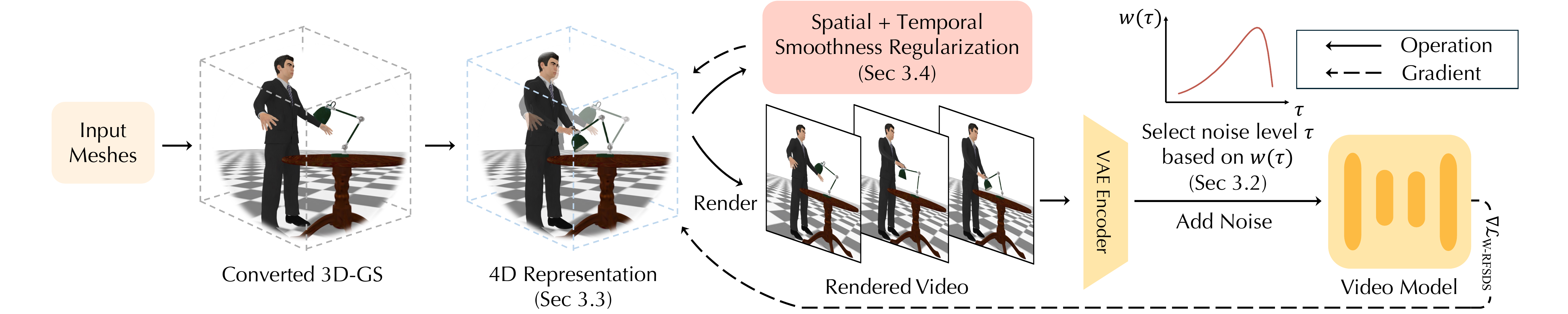}
\caption{
\textbf{Overview.} For the input meshes of a given scene, we first convert them into 3D-GS representations to enable smooth gradient computation. The converted 3D-GS models are then used to initialize a 4D representation (Sec.~\ref{sec:4D-representation}). We iteratively refine this 4D representation by sampling camera poses at each iteration, rendering the corresponding videos, and passing them to the video generation model to obtain optimization gradients (Sec.~\ref{sec:supervision}). Additionally, we compute regularization terms (Sec.~\ref{sec:regularization}) to enforce spatial and temporal smoothness during the optimization process.
}
\vspace{-12pt}
\label{fig:overview}
\end{figure*}

Given a 3D scene containing multiple dynamic objects represented by their static 3D snapshots, along with a text prompt describing how the scene should change over time (e.g., a man facing a lamp with the prompt ``\textit{the man lowers the head of the lamp with his hand}''), our goal is to generate a sequence of temporal deformations that drive the objects so that the resulting 3D animations aligns with the prompt.

Figure~\ref{fig:overview} shows an overview of our method. We  iteratively optimize a 4D scene motion representation using guidance signals distilled from a video generative model. In the following section, we detail the three main components in this framework: a strategy for distillation from modern rectified flow-based video generative models (Sec.~\ref{sec:supervision}), a robust and general 4D scene motion representation (Sec.~\ref{sec:4D-representation}), and regularization terms to ensure stable optimization (Sec.~\ref{sec:regularization}).

\subsection{Preliminary: Score Distillation Sampling}

The Score Distillation Sampling method \cite{poole2023dreamfusion} was introduced to distill 3D assets from image diffusion models \cite{ho2020denoising}. At each iteration, an image $\mathbf{z}$ is rendered from the 3D asset parameterized by $\theta$. Gaussian noise $\epsilon$ is then added to produce a noisy image $\mathbf{z}_\tau$, where the noise level $\tau$ is uniformly sampled from $(0,1)$. The noisy image $\mathbf{z}_\tau$ is subsequently fed into a image diffusion model, which predicts noise $\hat\epsilon$. SDS updates $\theta$ with the following gradient:
\begin{equation}
\begin{aligned}
&\nabla_\theta \mathcal{L}_\text{SDS} (\theta; \mathbf{z}, \mathbf{y} ) = \mathbb{E}_{\tau, \epsilon} \left[
w(\tau) \left( \hat\epsilon \left( \mathbf{z}_\tau; \tau, \mathbf{y} \right) - \epsilon \right)
\frac{\partial \mathbf{z}}{\partial \theta}
\right],
\end{aligned}
\label{eq:imgsds}
\end{equation}
where $w(\tau)$ is a weighting function.

Extending this idea to 4D generation follows the same principle: at each iteration, a video is rendered from the 4D asset, blended with noise, and then passed through the diffusion model, which provides gradients to update the 4D representation.

\subsection{Distilling from Rectified Flow Models}
\label{sec:supervision}

The above-mentioned 4D SDS algorithm is conceptually simple, yet it is non-trivial to apply them to distill from modern video generative models. The major obstacle is the gap between the diffusion architecture used in the original SDS target and the Rectified Flow (RF)-based model architecture in modern video generative models, such as Wan~2.2~\cite{wan2025} used in our paper.

To mitigate this architectural gap, we derive a novel SDS target for RF models. 
Similar to the derivation of SDS gradients for diffusion models~\cite{poole2023dreamfusion}, we align the optimization objective with the model’s training loss and express the SDS update rule for RF models as:

\begin{equation}
\begin{aligned}
&\nabla_\theta \mathcal{L}_\text{RFSDS} (\theta; z, \mathbf{y} ) = \\
&\mathbb{E}_{\tau, \epsilon} \left[
w(\tau) \left( \hat v \left( \mathbf{z}_\tau; \tau, \mathbf{y} \right) - \epsilon +\mathbf{z} \right) 
\frac{\partial \mathbf{z} }{\partial \theta} 
\right],
\end{aligned}
\label{eq:rfsds}
\end{equation}
where $\tau$ is the noise level uniformly sampled from $(0,1)$, $w(\tau)$ is the corresponding weight in the training schedule, $\epsilon$ is the added noise, $\mathbf{z}_\tau = (1-\tau)\mathbf{z} + \tau\epsilon$ denotes the noisy video, and $\hat v \left( \mathbf{z}_\tau; \tau, \mathbf{y} \right)$ is the predicted velocity.

A domain-specific noise sampling strategy is critical for this target to work well on our objective of optimizing scene deformations. We observed that the deformations are prone to be generated at higher noise levels $\tau$, as significant changes only happen when substantial noise is added. 
Based on this observation and the properties of $w(\tau)$, instead of sampling $\tau$ uniformly, we perform sampling according to a probability density function $\hat{w}(\tau) = \frac{1}{\int_{-\infty}^{\infty} w(\tau) \, \mathrm{d}\tau} w(\tau)$, which is the normalized form of $w(\tau)$. 

With this modification in sampling strategy, the weighted RFSDS update rule becomes:
\begin{equation}
\begin{aligned}
&\nabla_\theta \mathcal{L}_\text{W-RFSDS} (\theta; z, \mathbf{y} ) = \\
&\mathbb{E}_{\tau\sim\hat{w}(\tau), \epsilon} \left[
\left( \hat v \left( \mathbf{z}_\tau; \tau, \mathbf{y} \right) - \epsilon +\mathbf{z} \right) 
\frac{\partial \mathbf{z} }{\partial \theta} 
\right],
\end{aligned}
\label{eq:rfsds_ours}
\end{equation}

where the weighting term in RFSDS gradients defined in \cref{eq:rfsds} is eliminated to ensures the invariance of the expectation of gradients. Empirically, this yields more realistic generated motion, as shown in \cref{ssec:ablation}.

Practically, this noise sampling strategy is implemented with an annealing noise schedule~\cite{huangdreamtime,tangdreamgaussian} during the optimization. At each optimization step $i$ out of entire $I$ iterations, we set $\tau$ to be a fixed noise level $\tau_i$, which is obtained by solving:
\begin{equation}
\begin{aligned}
h(\tau_i) = 1 - \frac{i}{I + 1},
\label{eq:noise_schedule}
\end{aligned}
\end{equation}
where $h(\tau) = \int_{-\infty}^{\tau} \hat{w}(t) \, \mathrm{d}t$ is the cumulative distribution function (CDF) of $\hat{w}(\tau)$. This creates an annealing schedule in which $\tau$ gradually decreases over training, enabling coarse motion to form early and allowing fine deformations to be refined in later iterations.

\subsection{Hierarchical 4D Representation}
\label{sec:4D-representation}

Most existing 4D representations are highly unstable to optimize with the W-RFSDS target described above. Therefore, we introduce a hierarchical 4D representation that leverages natural locality of deformations in both spatial and temporal domain to stabilize the optimization process.

Our representation is composed of two components: a geometric representation of canonical shapes and a 4D motion representation that deforms the canonical geometry in different frames. The canonical shape of our 4D representation is represented with 3D-GS~\cite{kerbl20233d}.
Specifically, given $N$ mesh inputs, we convert them into 3D-GS models $\mathcal{S} = \{\mathcal{G}_i\}_{i=1}^{N}$ by optimizing directly on multi-view images rendered from the meshes, where each $\mathcal{G}_i$ denotes a converted 3D-GS model. 

At time $t$, the canonical 3D geometry is deformed with a set of deformation fields to represent the 4D motion of the 3D-GS scene $\mathcal{S}$. The set of deformation fields at time $t$ is denoted by $\mathcal{D}^t = \{\mathcal{T}_i^t\}_{i=1}^{N}$, where $\mathcal{T}_i^t$ denotes the deformation for object $i$ at time $t$. 

The 4D scene motion $\mathcal{D}$ is represented with a novel representation that injects hierarchical structures in both spatial and temporal domain, as detailed below.

\begin{figure}[t]
\includegraphics[width=\linewidth]{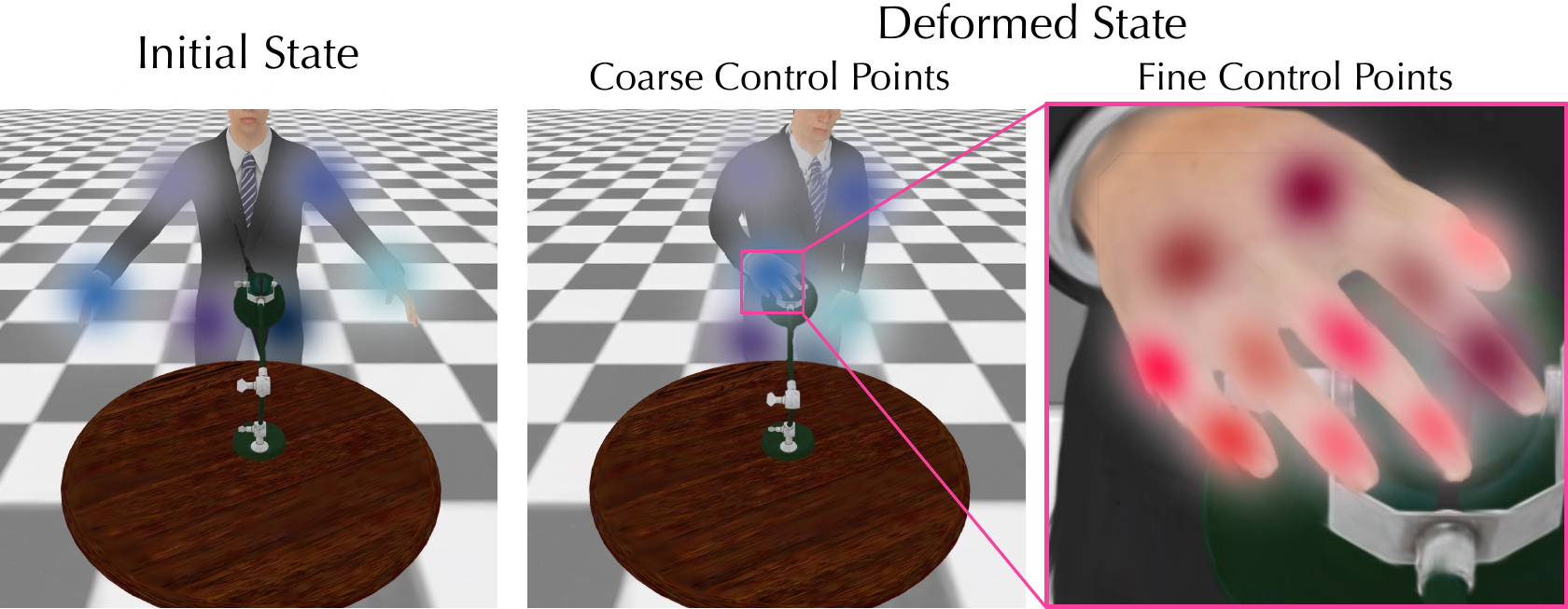}
\caption{
\textbf{Illustration of the hierarchical control point representation.} We represent the deformation using a spatial hierarchical structure. 
Coarse control points capture large-scale deformations, while fine control points refine local details.
} 
\vspace{-15pt}
\label{fig:hcp}
\end{figure}

\myparagraph{Spatial Hierarchy with Control Points.} The deformation fields $\mathcal{T}_i^t$ are spatially high-dimensional, and we reduce the dimensionality of this representation with a hierarchical control point-based representation.

Inspired by SC-GS~\cite{huang2024sc}, we represent $\mathcal{T}_i^t$ with a coarse level and a fine level of control points --- a sparse set of spatially-grounded blobs that controls a local spatial region of deformations. The coarse level of control points roughly dictates how an object will deform, and the fine level adds more details to the deformation. 

Specifically, each control point is defined by a mean $\boldsymbol{p}$ and a covariance matrix $\boldsymbol{\Sigma}$, which together determine its radius of influence. 
In addition, each control point maintains a sequence of deformations $(\mathbf{R}^t, \mathbf{T}^t)$ in $SE(3)$. 
The deformation of a Gaussian is obtained by blending transformations from neighboring control points using linear blend skinning. 
For a Gaussian $(\boldsymbol{\mu}, \boldsymbol{q}, \boldsymbol{S}, \mathcal{C}, o) \in \mathcal{G}_i$, we denote its $K$ nearest neighboring control points as $\mathcal{N}$. 
The deformed Gaussian at time $t$ is then computed as:
\begin{align}
    \label{eq:deform_gs_mean}
    \mu^t &= \sum\limits_{k \in \mathcal{N}} \beta_{k} \left(R_{k}^t(\mu - \mathbf{p}_k) + \mathbf{p}_k + T_{k}^t\right), \\
    \label{eq:deform_gs_rot}
    \mathbf{q}^t &= (\sum\limits_{k \in \mathcal{N}} \beta_{k} r_{k}^t) \otimes \mathbf{q},
\end{align}
where  $r_k^t \in \mathbb{R}^4$ are the quaternion representations of rotation on control point $k$, and $\otimes$ is the production of quaternions. Furthermore, $\beta_k$ in the formula denotes the blending weight of control point $k$, which is calculated through:
\begin{equation}
\label{eq:control_weight}
    \beta_{k} = \frac{\hat{\beta}_{k}}{\sum\limits_{l \in \mathcal{N}} \hat{\beta}_{l}} \text{, } \ \hat{\beta}_{k} = \text{exp} \left(-\frac{1}{2}\left[\left(\mu-p_{k}\right) \Sigma_{k}^{-1}\left(\mu-p_{k}\right)^{T}\right]  \right). 
\end{equation}

We optimize the bi-level sets of control points in a coarse-to-fine manner, following the noise schedule defined in \cref{eq:noise_schedule}.
When $\tau$ is large during the optimization process, substantial motion can be generated; however, the SDS gradients produced at such noise levels are often noisy. Conversely, when $\tau$ is annealed to a lower value, the gradients become more stable but are less capable of producing substantial deformations. To accompany with the inherent nature of this optimization process, we only optimize the coarse level of control points at earlier iterations when $\tau$ is large, and we introduce the fine control points later, once $\tau$ becomes smaller, to append their residual deformations:
\begin{align}
    \label{eq:high_dof_deform_mean}
    \mu^t_{\text{final}} &= \Delta \mu^t + \mu_t, \\
    \label{eq:high_dof_deform_rot}
    \mathbf{q}^t_{\text{final}} &= \Delta \mathbf{q}^t \otimes \mathbf{q}^t
\end{align}
where $\Delta \boldsymbol{\mu}^t$ and $\Delta \mathbf{q}^t$ denote the residual deformations from the fine layer of control points, computed in the same manner as in \cref{eq:deform_gs_mean} and \cref{eq:deform_gs_rot}.

After training, the deformation learned with Gaussians can be directly transferred to deform meshes. Concretely, we deform the mesh vertices using \cref{eq:deform_gs_mean} by substituting the Gaussian means with the vertex positions.

\begin{figure}[t]
\includegraphics[width=\linewidth]{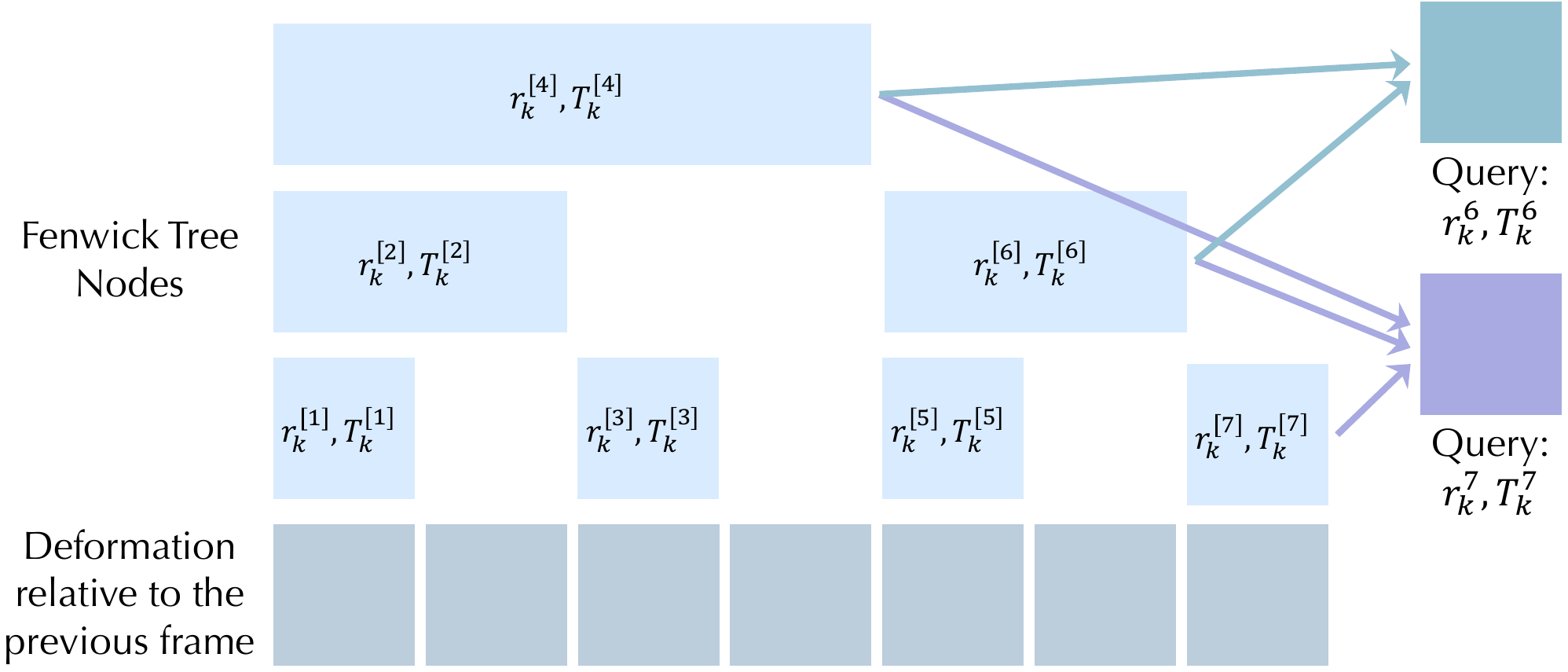}
\caption{
\textbf{Illustration of the Fenwick Tree representation.} Each node stores the cumulative deformation over a temporal range, allowing nearby frames to share parameters and naturally enforcing temporal coherence. For example, $(r_k^{[6]}, T_k^{[6]})$ encodes the accumulated deformation from frames 5–6. Queries for frames 6 and 7 then compose their deformations from a small, overlapping set of nodes, as shown in the figure.
} 
\vspace{-15pt}
\label{fig:fwtree}
\end{figure}

\myparagraph{Temporal Hierarchy with the Fenwick Tree.}
We further observe that deformations of later frames are challenging to learn if $(R^t, T^t)$ of frame $t$ are modeled independently from other frames. This can be explained by the fact that all deformations are initially initialized as zero vectors and the parameters of the first frame are kept frozen, leading to the significant deviation of deformations in later frames. 

To alleviate this issue, we represent the sequence of deformations for each control point $(R^t, T^t)$ with the Fenwick tree, a hierarchical data structure from theoretical algorithm design~\cite{fenwick1994new}. 
As illustrated in \cref{fig:fwtree}, for each control point $k$, we maintain nodes $\mathcal{F}_k = \{(r_k^{[j]}, T_k^{[j]})\}_{j=1}^{T}$, where each node encodes the accumulated deformation over a specific range of frames. This range-based decomposition allows deformations at different frames to share parameters through overlapping intervals, greatly improving temporal coherence and enable the learning of long-horizon motion.

The final deformation at frame $t$ is obtained by composing all relevant nodes:
\begin{align}
    \label{eq:fw_tree_rot}
     T_k^t &= \sum_{j \in \text{BIT}(t)} T_k^{[j]},\\
    \label{eq:fw_tree_trans}
    r_k^t &= \text{norm}(\sum_{j \in \text{BIT}(t)} r_k^{[j]}),
\end{align}
where $\text{BIT}(t)$ denotes the set of active nodes returned by the Fenwick query operation, and $\text{norm}(\cdot)$ ensures that the summed result forms a valid quaternion.

\subsection{Regularization}
\label{sec:regularization}
We introduce two regularization terms to further stabilize the optimization process: a temporal regularization loss to enforce smoothness over time and a spatial regularization loss to encourage local spatial consistency.

\myparagraph{Temporal Regularization.} When rendering the RGB video for computing the SDS gradients, we additionally render a 3D flow map video $\mathbf{F}$ from the same viewpoint, which is used for temporal regularization. To produce the flow map at frame $t$, we replace the color attribute of Gaussians in the 3D-GS rendering equation with $\boldsymbol{\mu}_i^t - \boldsymbol{\mu}_i^{t+1}$, where $\boldsymbol{\mu}_i^t$ denotes the mean of Gaussian $i$ at time $t$. After obtaining $\mathbf{F}$, the temporal regularization loss is defined as:
\begin{equation}
     \mathcal{L}_{\text{temp}} = \sum_t\sum_\textbf{p}||F_\textbf{p}^t||_2^2 ,
\end{equation}
where the inner summation is over all pixels $\mathbf{p}$, and $\mathbf{F}_{\mathbf{p}}^t$ represents the rendered 3D flow at pixel $\mathbf{p}$ and time $t$.

\myparagraph{Spatial Regularization.} To ensure spatially uniform regularization, we generate a uniformly distributed point cloud near the surface of each object $i$, deform it using the learned motion, and compute an As-Rigid-As-Possible (ARAP) loss \cite{sorkine2007rigid} over the resulting sequence of deformed point clouds. Specifically, we first compute a signed distance field (SDF) $\phi_i(\mathbf{x})$ from the mesh of object $i$. We then extract voxel centers near the surface as $\mathcal{S}_i=\{\mathbf{x} \mid |\phi_i(\mathbf{x})|\leq \tau, \textbf{x}\in V_s \}$,
where $V_s$ is the set of voxel centers on a grid with voxel size $s$, and $\tau$ is a predefined threshold. At each iteration, for every $\mathbf{x} \in \mathcal{S}_i$ and timestamp $t$, we compute its deformed position $\mathbf{x}^t$ using \cref{eq:deform_gs_mean} (with $\mu$ replaced by $\mathbf{x}$), thereby producing the deformed point set $\mathcal{S}_i^t=\{\mathbf{x}^t \mid \mathbf{x} \in \mathcal{S}_i \}$.
ARAP loss is then calculated as:
\begin{equation}
\begin{aligned}
     \mathcal{L}_{\text{ARAP}} = 
     \sum_{i, t, {\textbf{x}\in \mathcal{S}_i}, \textbf{y}\in \mathcal{N}_{\textbf{x}}} 
     ||\textbf{x}-\textbf{y}-\hat R_\textbf{x}(\textbf{x}^t-\textbf{y}^t)||_2^2,
\end{aligned}
\end{equation}
where $\mathcal{N}_{\mathbf{x}}$ denotes the set of the 10 nearest neighbors of $\mathbf{x}$ in $\mathcal{S}_i$, and $\hat{R}_{\mathbf{x}}$ is the estimated local rotation matrix at $\mathbf{x}$.

\begin{figure*}[t!]
\includegraphics[width=\linewidth]{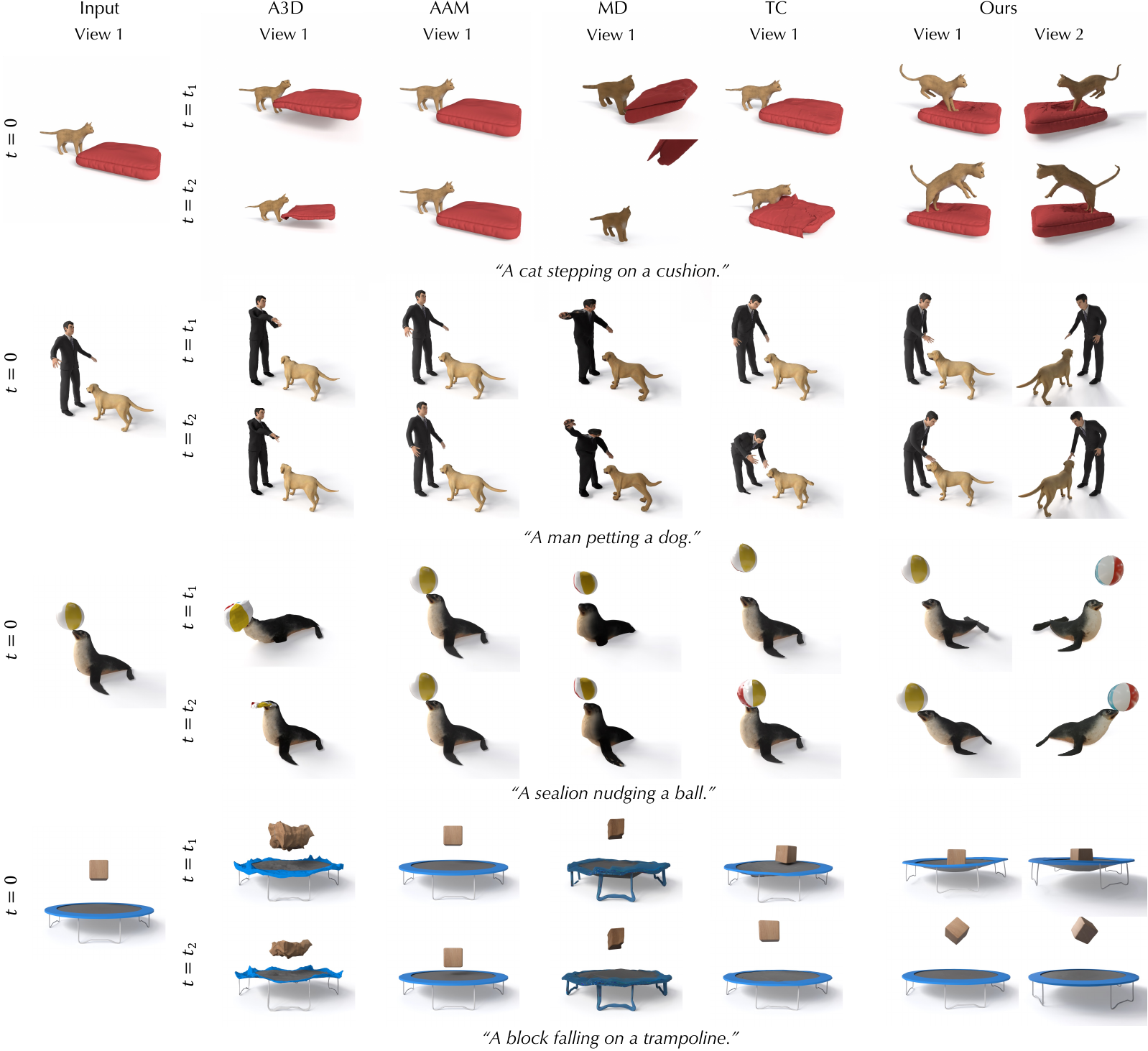}
\vspace{-18pt}
\caption{
\textbf{Qualitative comparisons.} We compare our approach with several mesh animation methods. Our method produces results that better align with the given prompts and exhibit more natural motion. In the figure, A3D refers to Animate3D~\cite{jiang2024animate3d}, AAM denotes AnimateAnyMesh~\cite{wu2025animateanymesh}, MD represents MotionDreamer~\cite{uzolas2025motiondreamer}, and TC corresponds to 4D reconstruction results from videos generated by TrajectoryCrafter~\cite{Yu_2025_ICCV}. For additional comparisons and full animation results, please refer to our supplementary website.
}
\vspace{-10pt}
\label{fig:qualitive_results}
\end{figure*}
\section{Experiments}

We evaluate our proposed method on a diverse dynamic scenes featuring multiple interacting objects. We compare our approach with several state-of-the-art baselines, each representing a distinct category of methods.

\subsection{4D Scene Motion Generation}

We compare our method against state-of-the-art mesh animation approaches, as well as
4D reconstructions from camera-controlled video models.
Specifically, we compare our approach with four baselines: Animate3D \cite{jiang2024animate3d}, AnimateAnyMesh \cite{wu2025animateanymesh}, MotionDreamer \cite{uzolas2025motiondreamer}, and TrajectoryCrafter \cite{Yu_2025_ICCV}. Animate3D generates multi-view videos using a multi-view video diffusion model and then performs 4D reconstruction on them. AnimateAnyMesh directly predicts mesh deformations using a pretrained Rectified Flow model. MotionDreamer first generates a video conditioned on the text prompt and a rendering of the given mesh, and then animates the mesh by performing diffusion feature matching with the generated video. 
We present results from our reimplementation using Wan 2.2, and provide results obtained with DynamiCrafter \cite{xing2024dynamicrafter} which was used in its original pipeline in the supplementary materials. TrajectoryCrafter is a video generation model that redirects
camera trajectories for monocular videos. We first generate a video using Wan 2.2, then produce corresponding multi-view videos with TrajectoryCrafter, and finally perform 4D reconstruction on the sampled videos. 

We select six scenes spanning diverse object categories for comparison: 
``A man petting a dog'', ``A cat stepping on a cushion'', ``A sealion nudging a ball'', ``A block falling on a trampoline'', ``Two men shaking hands'', and ``A robot picking up a block''. 
We additionally include comparisons between our method and baseline approaches for \textbf{single-object mesh animation} in the supplementary materials.

\myparagraph{Qualitative Comparisons.} Part of the qualitative results are shown in Figure~\ref{fig:qualitive_results}; please refer to the supplementary materials for the complete set of results. Our method exhibits stronger prompt alignment and generates more natural motion compared to existing approaches. Animate3D and AnimateAnyMesh fail to generate results that align with the given prompts, as they have not been extensively trained on 4D data containing multiple objects. MotionDreamer suffers from severe artifacts due to errors in diffusion feature matching when fitting meshes. Although 4D reconstruction from videos sampled via TrajectoryCrafter yields motions that follow the prompts, the results suffer from strong temporal inconsistencies and unnatural dynamics due to discrepancies among videos generated under different camera trajectories. This highlights the necessity of distilling a video model in our method.

\begin{table}[]
\centering
\caption{\textbf{Quantitative comparisons with baselines.} We conduct a user study on six scene animations to evaluate the performance. Additionally, we report the Semantic Adherence (SA) and Physical Commonsense (PC) metrics computed with VideoPhy-2~\cite{bansal2025videophy}. 
}
\vspace{-6pt}
\label{tab:quantitative_results}
\resizebox{\linewidth}{!}{
\begin{tabular}{lcccc}
\toprule
 & \multicolumn{2}{c}{User Study} & \multicolumn{2}{c}{VideoPhy-2} \\
\cmidrule(lr){2-3}\cmidrule(lr){4-5}
 & Alignment \(\uparrow\) & Realism \(\uparrow\)   & SA \(\uparrow\) & PC \(\uparrow\) \\
\midrule
Animate3D ~ & 0.34\%  & 0.51\% & 3.83 & 3.42 \\
AnimateAnyMesh ~  & 1.01\% & 0.51\% & 3.5 & \textbf{4.5} \\
MotionDreamer (DC) ~  & 0.51\% & 0.84\% & 3.42 & 4.08 \\
MotionDreamer (Wan) ~  & 0.84\% & 0.34\% & 3.5 & 3.83 \\
TrajectoryCrafter ~  & 9.60\% & 10.44\% & \underline{4.17} & 3.83 
\\
\midrule
\method (Ours) ~  & \textbf{87.71\%} & \textbf{87.37\%} & \textbf{4.33} & \underline{4.25} \\
\bottomrule
\end{tabular}}
\vspace{-15pt}
\end{table}

\myparagraph{Quantitative Comparisons.}
We perform a user study with 99 participants to compare the quality of our method with the baselines. Additionally, we utilize VideoPhy-2 \cite{bansal2025videophy} to automatically evaluate the rendered videos from two aspects: Semantic Adherence (SA) and Physical Commonsense (PC). As shown in Table~\ref{tab:quantitative_results}, our method achieves the highest score in SA and the second-highest score in PC. Note that AnimateAnyMesh achieves the highest Physical Commonsense (PC) score due to its common failure mode, where objects remain static—an outcome that aligns with physical commonsense but fails to follow the given prompt.

\begin{figure}[t]%
  \centering
  \includegraphics[width=\linewidth]{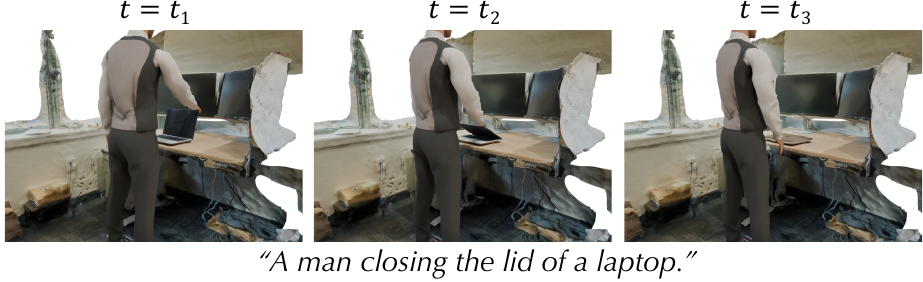}
  \vspace{-20pt}
  \caption{\textbf{Real-world object animation results.}}
  \vspace{-18pt}
  \label{fig:realworld_demo}
\end{figure}

\subsection{Extensions and Applications}
Beyond generating multi-object 4D motion, our framework naturally supports several extension and downstream uses.

\begin{figure*}[t!]
\centering
\includegraphics[width=\linewidth]{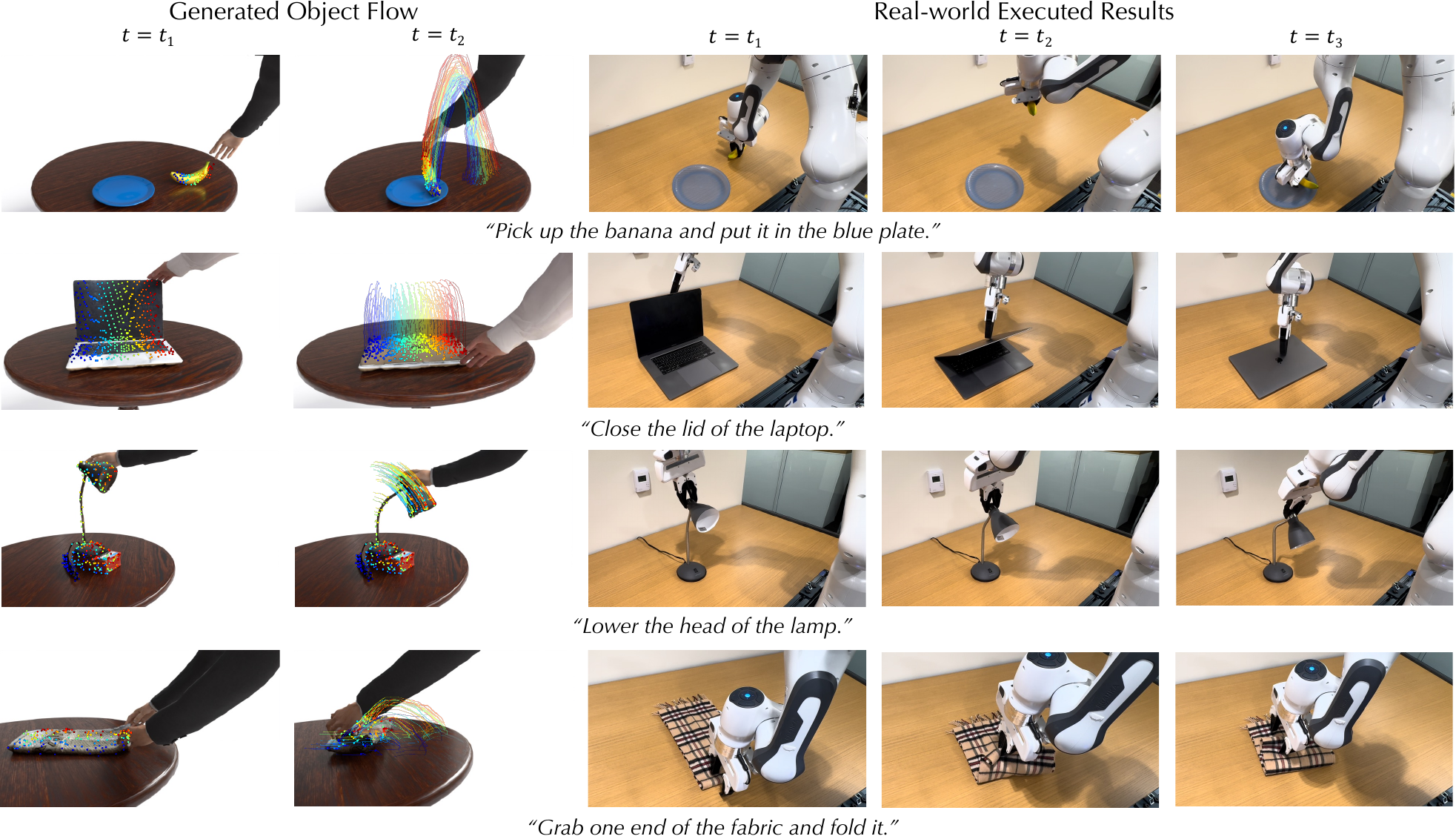}
\vspace{-15pt}
\caption{
\textbf{Robot manipulation guided by our generated dense object flow.} Given our generated dense object flow, the robot either grasps or pushes the object of interest in a manner that matches the flow. This allows effective manipulation of rigid objects (first row), articulated objects (second row), and deformable objects (third and fourth rows).
}
\vspace{-10pt}
\label{fig:robot_demo}\end{figure*}

\myparagraph{Long-Horizon Motion Generation.}
By using the last frame of the generated deformation as the input state for the subsequent generation process, we can extend our method to produce longer motion sequences. In Figure~\ref{fig:teaser}, we show an example motion sequence consisting of four actions.

\myparagraph{Real-world Object Animation.}
Since our method distills a video generative model trained extensively on real-world video data, it is robust and can be applied to animate scanned real-world objects without concern for the gap between synthetic and real-world data, as shown in Figure~\ref{fig:realworld_demo}.

\myparagraph{Robot Manipulation.}
We demonstrate that the dense object flow generated by our method can be utilized as guidance for manipulation of rigid, articulated, and deformable objects, as shown in Figure~\ref{fig:robot_demo}. We first use an off-the-shelf grasp planner~\cite{fang2023anygrasp} to propose a grasp on the relevant object. Then, the robot either grasps the object or moves to a pose for pushing the object, which is at an offset from the proposed grasp. Constrained by a rigid attachment forward model, where relative transformations of the end-effector also apply to the initial points on the object, a motion planner~\cite{kim2025pyroki} solves for a sequence of end-effector poses to minimize an objective consisting of transformed points to dense flow alignment, reachability, and pose smoothness costs.

\subsection{Ablation Studies}

\label{ssec:ablation}
\begin{figure}[t!]
\includegraphics[width=\linewidth]
{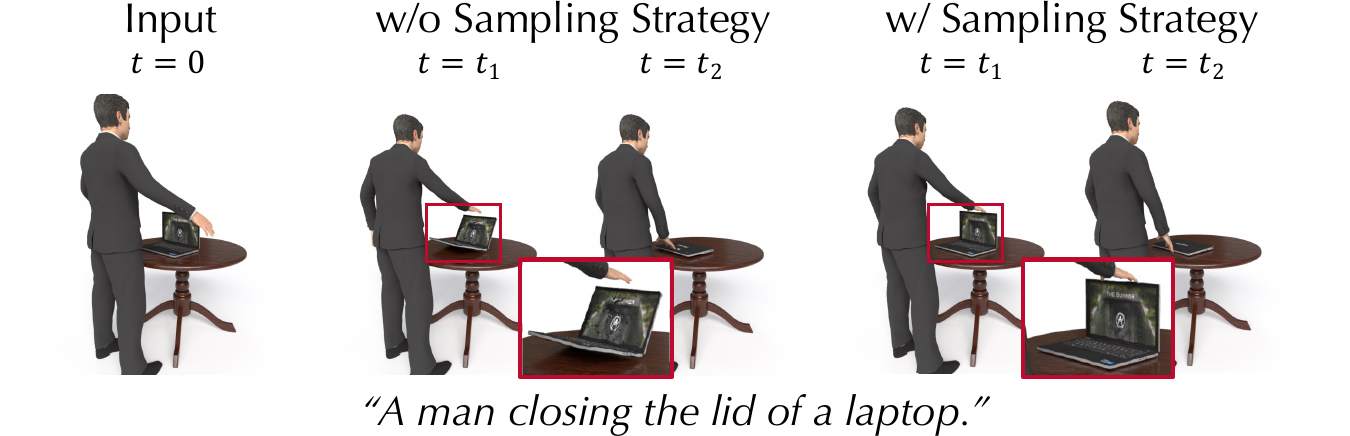}
\vspace{-18pt}
\caption{
\textbf{Ablation on noise-level sampling strategy.} 
Removing our noise-level sampling strategy leads to unnatural motion, such as the laptop appearing to float.
} 
\vspace{-5pt}
\label{fig:ablation_rs}
\end{figure}

\myparagraph{Noise Level Sampling Strategy}
We compare the effectiveness of the noise-level sampling strategy (Sec.~\ref{sec:supervision}) against uniform noise sampling with weighting. As shown in Figure~\ref{fig:ablation_rs}, unrealistic results emerge under uniform sampling due to insufficient coverage of noise levels that inject motion. In this case, the laptop appears to float above the table.

\begin{figure}[t!]
\includegraphics[width=\linewidth]{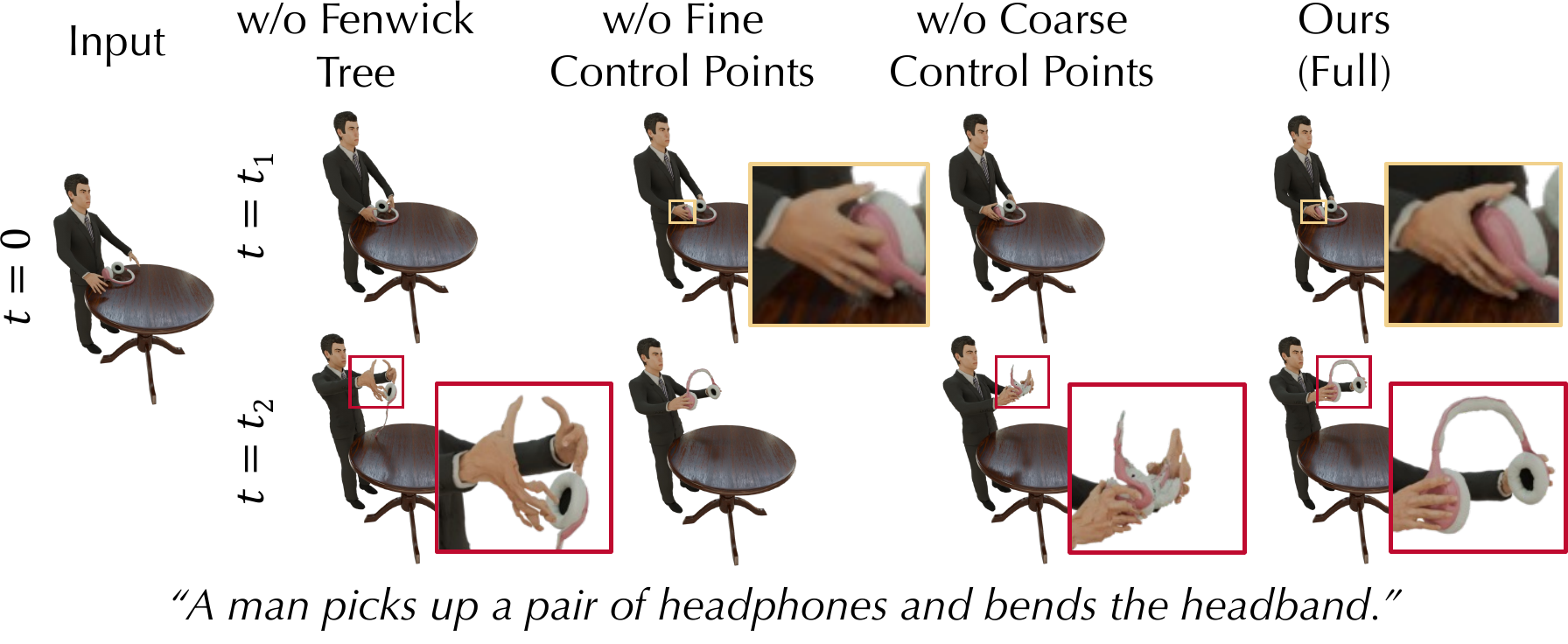}
\vspace{-18pt}
\caption{
\textbf{Ablations on components in the 4D representation.} Removing the Fenwick Tree leads to severe artifacts in later frames; removing fine control points prevents detailed deformation; and removing coarse control points causes distortions.
} 
\vspace{-18pt}
\label{fig:ablation_4dr}
\end{figure}

\myparagraph{4D Representation.}
We study two key components of our 4D representation: the Fenwick tree for modeling deformation sequences and the hierarchical control-point structure. Results are shown in Figure~\ref{fig:ablation_4dr}. Removing the Fenwick tree leads to noticeable artifacts, as later frames become extremely difficult to learn when each deformation is modeled independently. Removing the fine control-point layer prevents the model from producing detailed motions (e.g., grasping). Conversely, starting with the fine layer from the beginning also introduces artifacts, since the noise injected at early iterations cannot be effectively smoothed without an initial coarse stage.

\begin{figure}[t!]
\includegraphics[width=\linewidth]{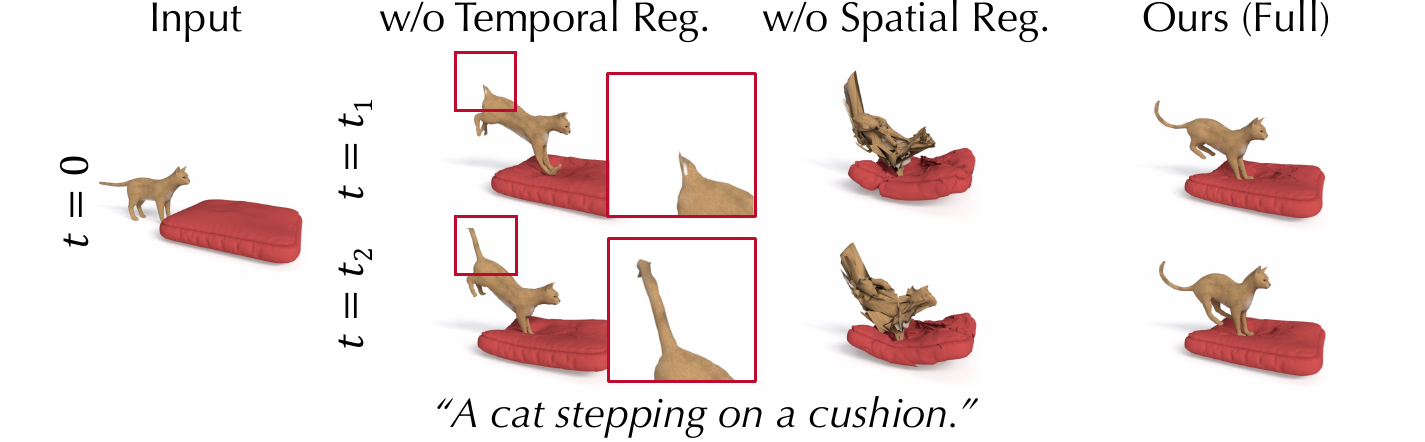}
\vspace{-15pt}
\caption{
\textbf{Ablations on regularization losses.}
Removing temporal regularization results in flickering, while removing spatial regularization results in distortions.
} 
\vspace{-15pt}
\label{fig:ablation_rg}
\end{figure}

\myparagraph{Regularization.}
We show that the regularization losses are necessary. Removing them results in temporal flickering (e.g., the tail suddenly appearing when temporal regularization is removed) and visual artifacts (when spatial regularization is removed), as shown in Figure~\ref{fig:ablation_rg}.

\section{Conclusion}

We introduce a robust, scalable, and versatile approach to generate scene-level 4D object motion given only 3D shapes as input. Our pipeline works effectively for diverse natural phenomena and opens new possibilities of scalable 4D generation with guidance from video generative models. It also enables downstream applications as we demonstrated in the robotics manipulation.

{
    \small
    \bibliographystyle{ieeenat_fullname}
    \bibliography{main}
}

\appendix
\clearpage
\onecolumn
\section{Implementation Details}

\subsection{Pipeline Implementation Details}
The 3D assets used in our experiments are downloaded from Sketchfab~\cite{sketchfab} and BlenderKit~\cite{blenderkit}, and we construct the static scene snapshots in Blender~\cite{blender}. Rendering of 3D-GS~\cite{kerbl20233d} for both mesh-based initialization and 4D optimization is performed using \texttt{gsplat}~\cite{ye2025gsplat}. We adopt the Wan 2.2 (14B) image-to-video model \cite{wan2025} as our video generation model. All training is conducted at a resolution of $832\times464$ (the default for Wan~2.2), and deformation sequences of $41$ frames are optimized.

Control points are initialized based on the center points of an occupancy grid. Specifically, for each object, we first compute a signed distance field (SDF) $\phi_i(\mathbf{x})$ from its given mesh. We then extract the set of voxel centers within the object as $\mathcal{I}_i=\{\mathbf{x} \mid \phi_i(\mathbf{x})\leq 0, \textbf{x}\in V_s \}$, where $V_s$ denotes all voxel center points in a grid with voxel size $s$. Finally, we apply farthest point sampling followed by K-means clustering on $\mathcal{I}_i$ to determine the positions $\mathbf{p}_k$ of the control points. We further initialize the scale in each control point's covariance matrix $\mathbf{\Sigma}_k$ as the average distance to its three nearest neighboring control points, and set the initial rotation to the identity. For stable optimization, we keep $\mathbf{p}_k$ fixed and only optimize $\mathbf{\Sigma}_k$ during training. In the training of the deformations, we additionally introduce a split training schedule: at a iteration 100, we reinitialize all deformations after 30 to the deformation at 30, which further facilitates stable learning for later frames.

We use the log-linear learning rate schedule adopted in 3D-GS. The learning rate for the deformations stored in the Fenwick tree decays from $0.006$ to $0.00006$. The learning rate for the scales of the control points follows the same decay (from $0.006$ to $0.00006$), while the learning rate for rotations decays from $0.003$ to $0.00003$. The CFG \cite{ho2022classifier} scale is linearly decayed from 25 to 12.
The weight for the temporal regularization loss is decayed from $9.6$ to $1.6$, and the weight for the spatial regularization loss is decayed from $3000$ to $300$.
The voxel size $s$ used for extracting the uniformly distributed point cloud in temporal regularization and for initializing control points is automatically determined via binary search such that the number of voxel centers near the surface satisfies $|\mathcal{S}_i| \approx 7500$.
Each asset is trained for 2{,}000 iterations with a batch size of $4$, requiring approximately $20$ hours on an NVIDIA~H200 GPU.

\subsection{Robot Manipulation Implementation Details}
For the objects used to generate dense object flow, we directly scanned the real objects in the ``pick banana'' and ``lower lamp'' cases and fed the scans into our pipeline. For the other cases, due to challenges in accurately scanning the objects, we instead measured their length statistics and created digital cousins with matching dimensions in Blender before inputting them into our pipeline.

\subsection{Baseline Implementation Details}
For Animate3D~\cite{jiang2024animate3d} and AnimateAnyMesh~\cite{wu2025animateanymesh}, we merge all objects in the scene into a single mesh and directly input it into their pipelines. For MotionDreamer~\cite{uzolas2025motiondreamer}, we follow their setup and use Neural Jacobian Fields (NJF)~\cite{aigerman2022neural} as the animation model, training a separate NJF for each object. For robust 4D reconstruction of videos sampled from TrajectoryCrafter~\cite{Yu_2025_ICCV}, we use a coarse set of control points with a Fenwick-tree–based deformation sequence as the 4D representation. We additionally apply both temporal and spatial regularization losses during optimization.

\section{Derivation of SDS for Rectified Flow Models}

When sampling noise levels $\tau$ uniformly from $\mathcal{U}(0,1)$, the training loss of a Rectified Flow (RF) model~\cite{liu2023flow, esser2024scaling} is:
\begin{equation}
\mathcal{L}_\text{RF}(\theta; \mathbf{z}, \mathbf{y}) 
= \mathbb{E}_{\tau\sim\mathcal{U}(0,1),\, \epsilon} 
\left[
w(\tau)\,\big\| \hat v(\mathbf{z}_\tau; \tau, \mathbf{y}) - (\epsilon - \mathbf{z}) \big\|^2
\right],
\label{eq:rf_train_loss}
\end{equation}
where $\epsilon \sim \mathcal{N}(0, I)$ and $\mathbf{z}_\tau=(1-\tau)\mathbf{z} + \tau \epsilon$ is the linearly interpolated latent.

Taking the derivative of $\mathcal{L}_\text{RF}$ with respect to $\mathbf{z}$ yields:
\begin{equation}
\begin{aligned}
\nabla_{\mathbf{z}} \mathcal{L}_\text{RF}(\theta; \mathbf{z}, \mathbf{y})
= 
\mathbb{E}_{\tau\sim\mathcal{U}(0,1),\, \epsilon}
\left[
w(\tau)\,
\big( \hat v(\mathbf{z}_\tau; \tau, \mathbf{y}) - (\epsilon - \mathbf{z}) \big)
\left( \frac{\partial \hat v(\mathbf{z}_\tau; \tau, \mathbf{y})}{\partial \mathbf{z}} + I \right)
\right].
\end{aligned}
\label{eq:rf_grad_wrt_z}
\end{equation}

Following the derivation style of Score Distillation Sampling (SDS)~\cite{poole2023dreamfusion}, we omit the term that backpropagates through the RF model,
$\frac{\partial \hat v(\mathbf{z}_\tau; \tau, \mathbf{y})}{\partial \mathbf{z}}$, 
and apply the chain rule from $\mathbf{z}$ back to the 4D representation parameters $\theta$.
This gives the RF-SDS gradient used in the main text:
\begin{equation}
\nabla_\theta \mathcal{L}_\text{RFSDS}(\theta; \mathbf{z}, \mathbf{y})
=
\mathbb{E}_{\tau,\, \epsilon}
\left[
w(\tau)\,
\big( \hat v(\mathbf{z}_\tau; \tau, \mathbf{y}) - (\epsilon - \mathbf{z}) \big)\,
\frac{\partial \mathbf{z}}{\partial \theta}
\right].
\label{eq:rfsds}
\end{equation}

\section{More Experiment Results}

In this section, we present additional experimental results for our method.

\begin{figure*}[t!]
\includegraphics[width=\linewidth]{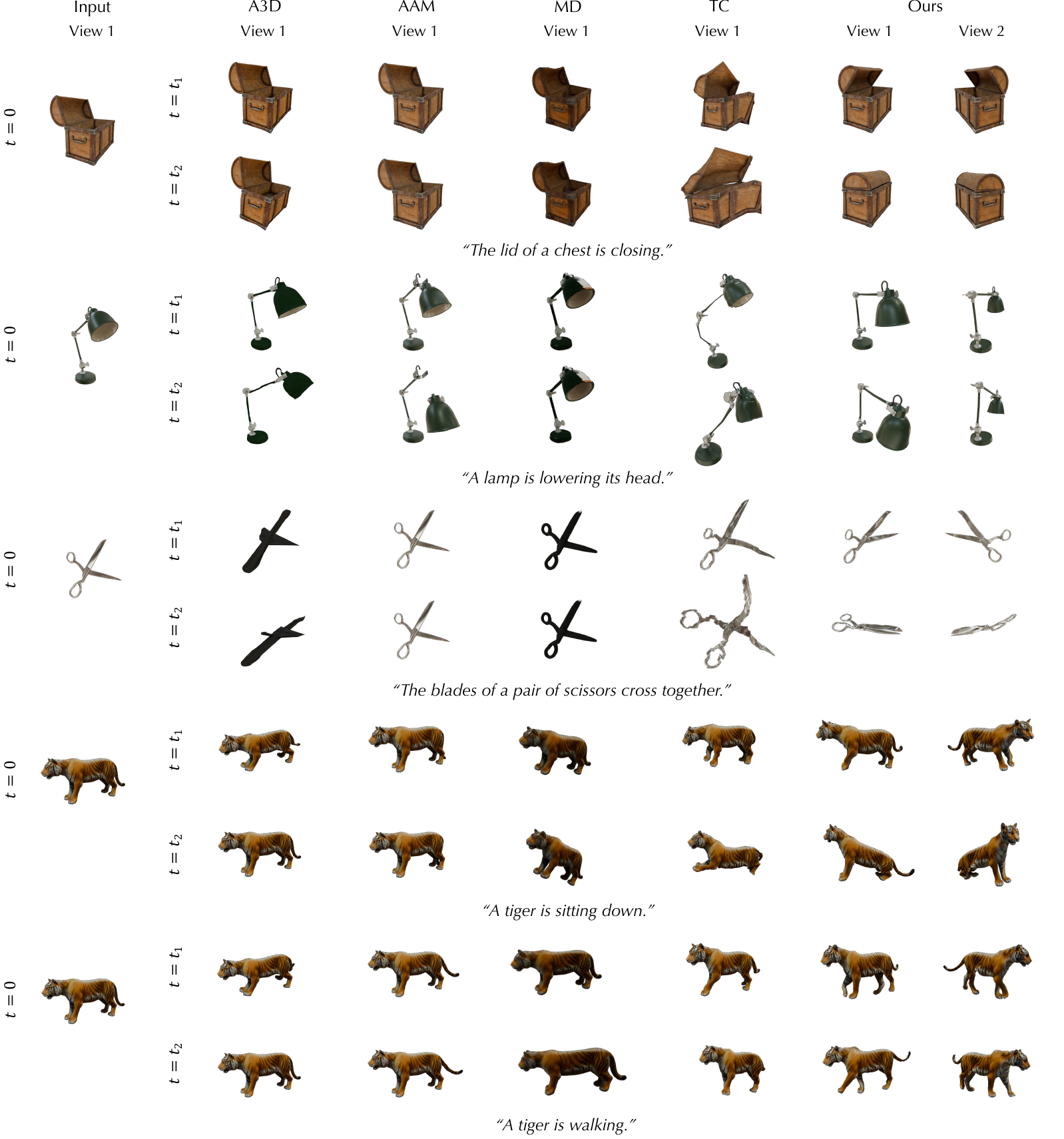}
\caption{
\textbf{Qualitative comparisons on single mesh animation.} We compare our approach with several mesh animation methods. Our method produces results that better align with the given prompts and exhibit more natural motion. In the figure, A3D refers to Animate3D~\cite{jiang2024animate3d}, AAM denotes AnimateAnyMesh~\cite{wu2025animateanymesh}, MD represents MotionDreamer~\cite{uzolas2025motiondreamer}, and TC corresponds to 4D reconstruction results from videos generated by TrajectoryCrafter~\cite{Yu_2025_ICCV}.
}
\label{fig:results_single}
\end{figure*}

\subsection{Comparison on Single Mesh Animation}

We further compare our method with baselines on the task of single-object mesh animation. The set of baselines follows the main paper: Animate3D~\cite{jiang2024animate3d}, AnimateAnyMesh~\cite{wu2025animateanymesh}, MotionDreamer~\cite{uzolas2025motiondreamer}, and 4D reconstruction from videos generated by TrajectoryCrafter~\cite{Yu_2025_ICCV}.
We evaluate all methods on five prompts: ``The lid of a chest is closing", ``A lamp is lowering its head", ``The blades of a pair of scissors cross together", ``A tiger is sitting down", ``A tiger is walking".

Qualitative results are shown in Figure~\ref{fig:results_single}. Our method consistently achieves better prompt alignment and produces more natural motion than existing approaches. For quantitative evaluation, we conducted a user study with 50 participants comparing our results against all baselines: \textbf{89.6\%} of participants rated our method highest in prompt alignment, and \textbf{84\%} rated it highest in motion realism. These results further indicate the strength of our approach relative to existing methods. The full results are provided in Table~\ref{tab:full_single_object}.

\begin{table*}[t]
\centering
\resizebox{\textwidth}{!}{
\begin{tabular}{llrrrrrrr}
\toprule
\textbf{Metric} & \textbf{Object} & \textbf{Animate3D} & \textbf{AnimateAnyMesh} 
& \textbf{MotionDreamer (Orig)} & \textbf{MotionDreamer (Wan)} 
& \textbf{TC} & \textbf{Ours} & \textbf{Total} \\
\midrule

\multirow{8}{*}{\textbf{Prompt Alignment}}
& Cat      & 0 & 2 & 0 & 1 & 0  & 96 & 99 \\
& Dog      & 0 & 3 & 0 & 1 & 7  & 88 & 99 \\
& Hugging  & 1 & 0 & 0 & 0 & 1  & 97 & 99 \\
& Robot    & 0 & 0 & 1 & 1 & 2  & 95 & 99 \\
& Sea Lion & 1 & 0 & 1 & 2 & 43 & 52 & 99 \\
& Brick    & 0 & 1 & 1 & 0 & 4  & 93 & 99 \\
& \textbf{Avg} & 0.3333 & 1 & 0.5 & 0.8333 & 9.5 & 86.8333 & 99 \\
\midrule

\multirow{8}{*}{\textbf{Motion Realism}}
& Cat      & 0 & 0 & 1 & 1 & 2  & 95 & 99 \\
& Dog      & 1 & 2 & 1 & 0 & 11 & 84 & 99 \\
& Hugging  & 0 & 0 & 0 & 0 & 3  & 96 & 99 \\
& Robot    & 0 & 1 & 1 & 1 & 1  & 95 & 99 \\
& Sea Lion & 1 & 0 & 2 & 0 & 37 & 59 & 99 \\
& Brick    & 1 & 0 & 0 & 0 & 8  & 90 & 99 \\
& \textbf{Avg} & 0.5 & 0.5 & 0.8333 & 0.3333 & 10.3333 & 86.5 & 99 \\
\bottomrule
\end{tabular}
} %
\caption{Raw results of the user study on generating scene-level 4D motion. We show the number of vote from each participant on which option they consider the best under certain metric.}
\label{tab:full_multi_object}
\end{table*}

\begin{table*}[t]
\centering
\resizebox{\textwidth}{!}{
\begin{tabular}{llrrrrrrr}
\toprule
\textbf{Metric} & \textbf{Object} 
& \textbf{Animate3D} 
& \textbf{AnimateAnyMesh} 
& \textbf{MotionDreamer (Orig)} 
& \textbf{MotionDreamer (Wan)} 
& \textbf{TC} 
& \textbf{Ours} 
& \textbf{Total} \\
\midrule

\multirow{7}{*}{\textbf{Prompt Alignment}} 
& Chest      & 2 & 0 & 1 & 0 & 0 & 47 & 50 \\
& Lamp       & 0 & 1 & 2 & 1 & 0 & 46 & 50 \\
& Scissors   & 1 & 0 & 1 & 1 & 0 & 47 & 50 \\
& Sitting    & 4 & 1 & 1 & 0 & 5 & 39 & 50 \\
& Walking    & 1 & 0 & 3 & 0 & 1 & 45 & 50 \\

& \textbf{Avg (raw)} 
              & 1.6 & 0.4 & 1.6 & 0.4 & 1.2 & 44.8 & 50 \\

\midrule

\multirow{7}{*}{\textbf{Motion Realism}} 
& Chest      & 2 & 0 & 1 & 1 & 1 & 45 & 50 \\
& Lamp       & 5 & 0 & 2 & 0 & 0 & 43 & 50 \\
& Scissors   & 0 & 1 & 7 & 0 & 1 & 41 & 50 \\
& Sitting    & 5 & 1 & 2 & 0 & 6 & 36 & 50 \\
& Walking    & 2 & 2 & 1 & 0 & 0 & 45 & 50 \\

& \textbf{Avg (raw)} 
              & 2.8 & 0.8 & 2.6 & 0.2 & 1.6 & 42 & 50 \\

\bottomrule
\end{tabular}
}
\caption{User study results for quantitative comparison on single-object 4D motion generation. }
\label{tab:full_single_object}

\end{table*}

\subsection{Full Table for User Study}
In Table~\ref{tab:full_multi_object} and Table~\ref{tab:full_single_object}, we provide the complete user study results, including the number of participants who preferred each method for each scene. Across all scenes, our method receives the highest preference in both prompt alignment and motion realism.

\begin{figure*}[t!]
\includegraphics[width=\linewidth]{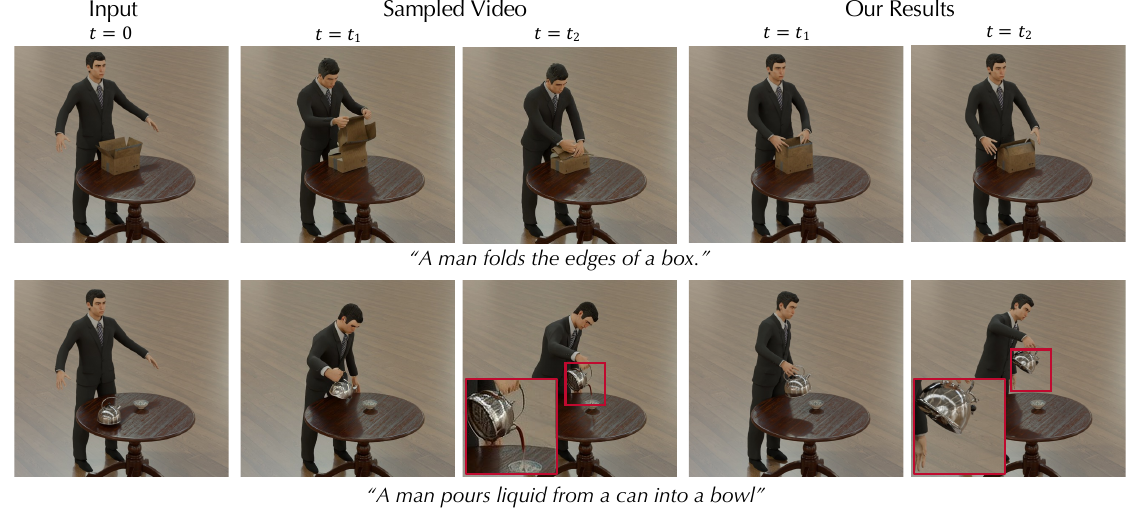}
\caption{
\textbf{Failure Cases.} The failure in the first row is due to limitations of the video generative model: it cannot produce motion that matches the prompt, as evidenced by its inability to sample videos aligned with the described action. The failure in the second row arises because our method cannot generate objects that were not present in the initial static scene. As a result, no liquid can appear when prompted, since the system cannot generate newly emerging objects.
}
\label{fig:failure}
\end{figure*}

\subsection{Failure Cases}
\label{sec:failure_case}

Our failure cases mainly arise from two factors: (1) limitations of the underlying video generative model, and (2) the inability to handle objects that do not exist in the static snapshot but appear later in the motion sequence. Examples are shown in Figure~\ref{fig:failure}. Our failure cases mainly arise from two factors: (1) limitations of the underlying video generative model, and (2) the inability to handle objects that do not exist in the static snapshot but appear later in the motion sequence. Examples are shown in Figure~\ref{fig:failure}. We elaborate on them below. 

\myparagraph{Video Generative Model Limitation.} Because our approach distills from a pretrained video generation model, its capabilities are inherently linked to those of the underlying model. If the generator cannot synthesize videos aligning with the prompt, our 4D optimization receives misleading gradients. In such cases, our method cannot generate the correct motion. This is shown in the first row of Figure~\ref{fig:failure}, where the video model repeatedly fails to sample videos consistent with the prompt, leading our method to produce incorrect motion.

\myparagraph{Inability to Handle Newly Appearing Objects.}
Another limitation of our method is that it cannot handle objects that do not exist in the initial static snapshot. Our 4D representation only deforms the geometry present at the start, so any object that should appear later in the sequence cannot be created. When the prompt involves new objects entering the scene, the supervision asks for motion that the system cannot produce. In these cases, the optimization either omits the requested effect or yields incomplete motion, as illustrated in the second row of Figure~\ref{fig:failure}, where no liquid appears because the system cannot introduce new geometry.

\section{Limitation and Future Work}
Although our method can generate dynamic scenes with highly realistic interactions among multiple objects, there remain several limitations that point to promising directions for future work. 
For the failure cases described in Sec.~\ref{sec:failure_case}, those arising from limitations of the underlying video generative model may be alleviated as video generation technology continues to improve. 
For failures caused by newly appearing objects that are not present in the initial static scene, a potential solution is to incorporate a module capable of generating new geometry during the optimization process.

Apart from the failure cases, another limitation of our method is its extensive training time. In our observations, a substantial portion of the runtime is spent backpropagating through the VAE \cite{kingma2014vae}. A promising future direction is to develop a distillation strategy that avoids backpropagating through the VAE entirely. This may be feasible because our objective is to generate motion rather than RGB appearance, suggesting that full VAE gradients may not be strictly necessary for effective motion supervision.

\begin{figure*}[t!]
\includegraphics[width=\linewidth]{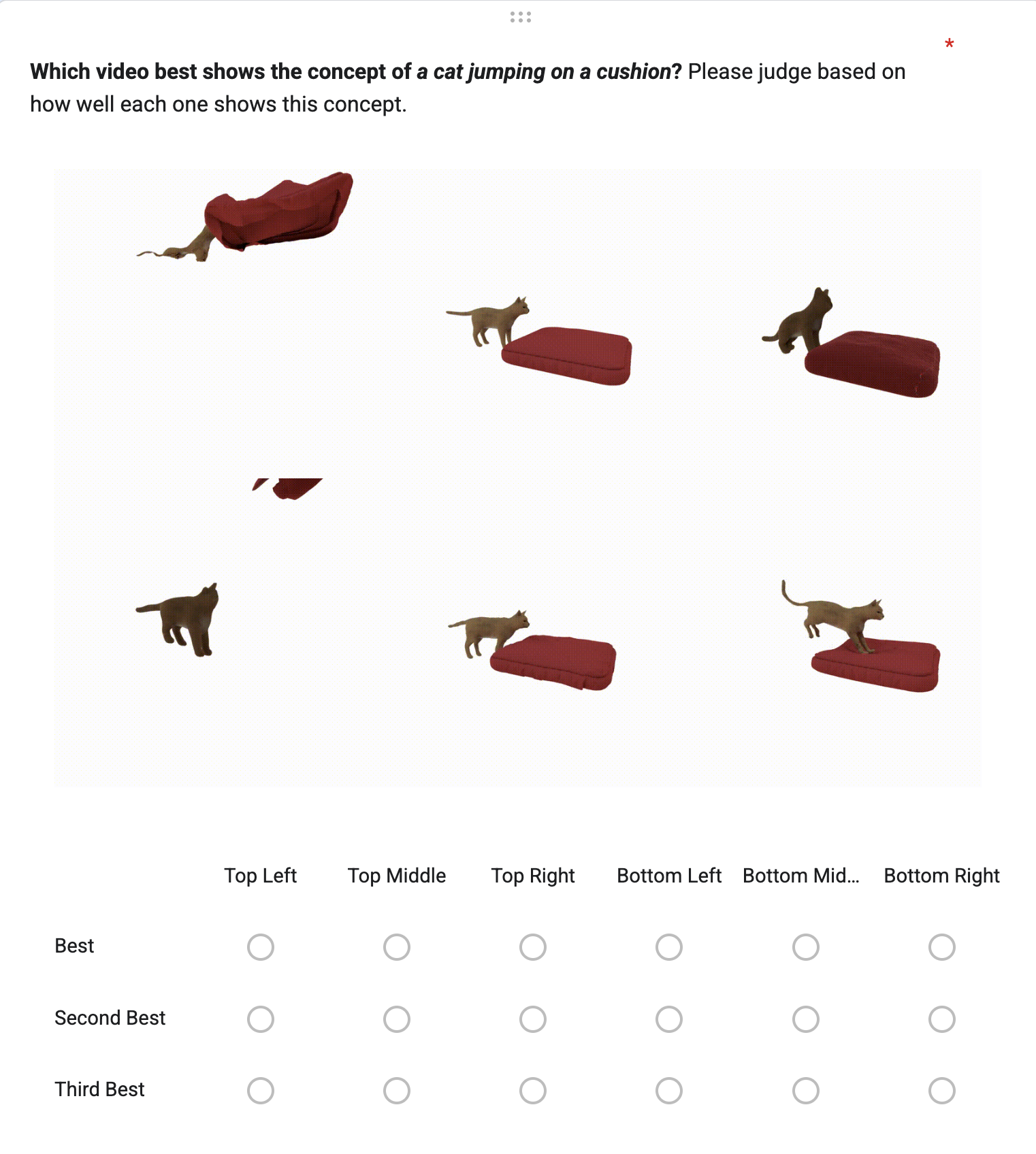}
\caption{
\textbf{Screenshot of the user study question on Prompt Alignment.}
}
\label{fig:user_study_1}
\end{figure*}

\begin{figure*}[t!]
\includegraphics[width=\linewidth]{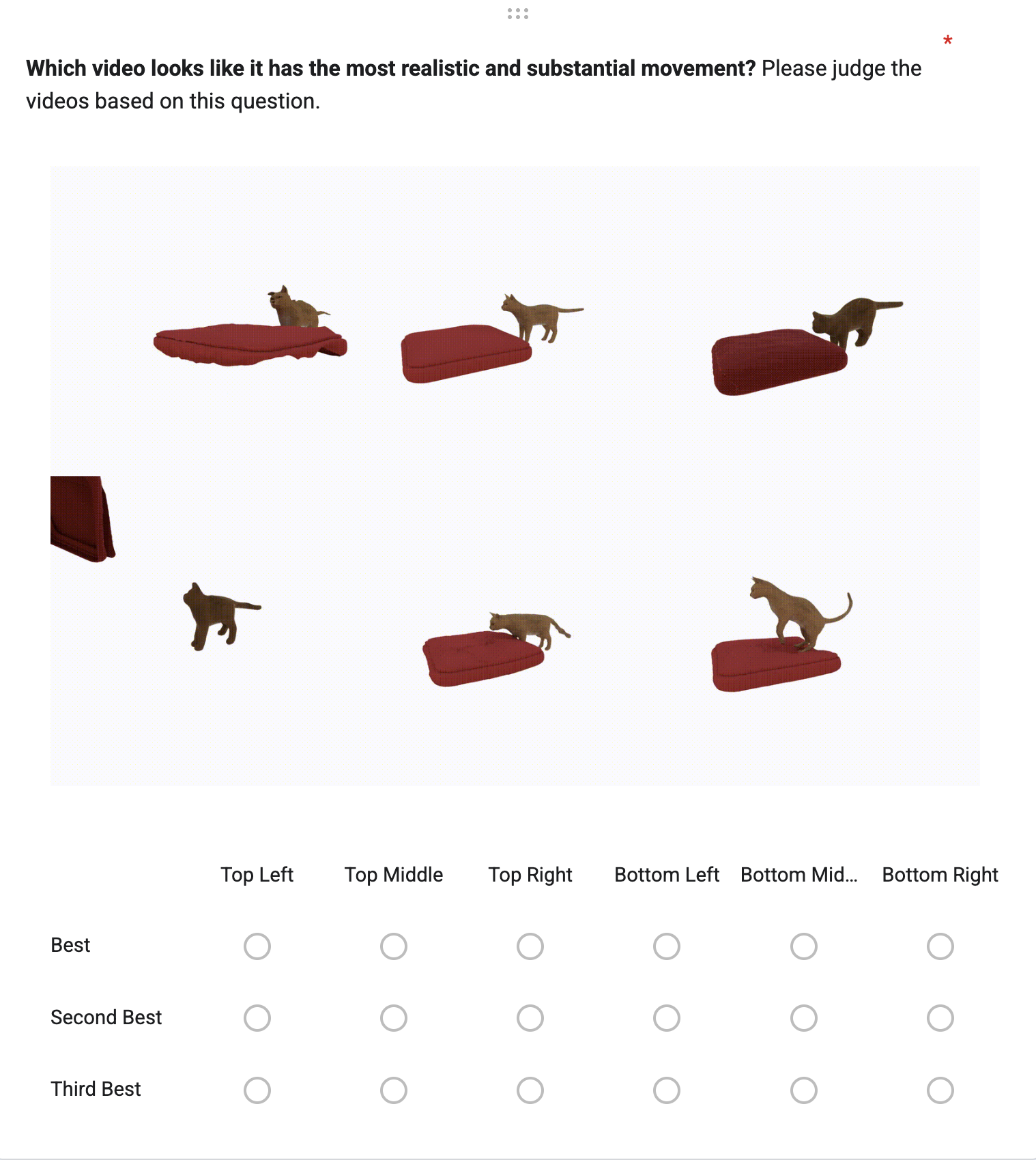}
\caption{
\textbf{Screenshot of the user study question on Motion Realism.}
}
\label{fig:user_study_2}
\end{figure*}

\section{User Study Template}
We provide screenshots of the user study interface in Figure~\ref{fig:user_study_1} and Figure~\ref{fig:user_study_2}. Participants were asked to select the best, second-best, and third-best results among five methods. From left to right and top to bottom, the corresponding methods are: Animate3D~\cite{jiang2024animate3d}, AnimateAnyMesh~\cite{wu2025animateanymesh}, MotionDreamer~\cite{uzolas2025motiondreamer} using DynamiCrafter~\cite{xing2024dynamicrafter}, MotionDreamer using Wan 2.2~\cite{wan2025}, 4D reconstruction from videos generated by TrajectoryCrafter~\cite{Yu_2025_ICCV}, and our method.

\end{document}